\documentclass[letterpaper, 10 pt, conference]{ieeeconf}
\IEEEoverridecommandlockouts
\AtBeginDocument{%
  }

\usepackage{pgfplots}
\usepackage{subcaption}
\usepackage{graphicx}
\usepackage{amsfonts}
\usepackage{algpseudocode}
\usepackage{algorithm}
\usepackage{amsmath}
\usepackage{lipsum}

\DeclareMathOperator*{\argmin}{arg\,min}

\usepackage{geometry}
\newtheorem{theorem}{Theorem}

\usepackage[backend=biber,style=ieee]{biblatex}
\DeclareFieldFormat{journaltitle}{#1}
\title{DARRMS - An Efficient Algorithm for Dynamic Attention Radius in Resource-Constrained Multi-Agent Systems}

\author{Benjamin Alcorn and Eman Hammad%
\thanks{B. Alcorn is with the Department of Electrical \& Computer Engineering, Texas A\&M University, 
        College Station, TX 77845, USA
        {\tt\small benjamin.alcorn@tamu.edu}}%
\thanks{E. Hammad is with the Departments of Engineering Technology, and Electrical \& Computer Engineering,
        Texas A\&M University, College Station, TX 77845, USA
        {\tt\small eman.hammad@tamu.edu}}%
}

\begin{document}

\maketitle
\thispagestyle{empty}
\pagestyle{empty}

\begin{abstract}
    Multi-agent systems are integral tools for various domains such as robotics, cybersecurity, and autonomous vehicle planning. These types of systems often have constraints on the computational resources, leading to a need for efficient lightweight algorithms. Traditional decision making frameworks often assume ideal conditions, such as full observability and unlimited computational capacity, which do not align with real-world challenges. In this paper, we introduce a new algorithm that allows for reduced demand on computational resources without a large cost of other performance metrics. Agents will limit their observability to some attention radius, which intentionally allows them to ignore parts of the environment that might be unnecessary for action planning. By optimizing both the attention radius and decision-making, our approach enhances coordination and scalability in uncertain environments. Through both theoretical analysis and empirical validation, we demonstrate the effectiveness of adaptive observation in improving system performance and maintaining robust decision-making strategies in resource-constrained systems. 
\end{abstract}

\section{Introduction}
\label{sec:intro}
    Multi-agent systems (MAS) are increasingly prevalent in domains such as robotics, economics, cybersecurity, and autonomous transportation, where multiple autonomous entities must interact, collaborate, or compete to achieve specific objectives. Effective decision-making in these systems is a complex challenge, as agents must account for uncertain environments, competing incentives, and limited computational or sensing resources. Traditional approaches to multi-agent decision-making often assume full observability and unlimited computational capacity, but real-world applications frequently operate under constraints that necessitate more adaptive and efficient strategies.

One well-established framework for strategic decision-making in multi-agent settings is the Stackelberg game, a hierarchical model in which a leader makes a decision while anticipating the responses of one or more followers. This framework is widely used in scenarios where agents have asymmetric roles and access to information, such as security games, traffic management, and economic markets. Stackelberg games offer a structured way to model agent interactions, allowing leaders to optimize their actions while accounting for the rational behavior of followers. 

A key consideration in resource-constrained MAS is the balance between observation and efficiency. In many cases, agents must decide how much information to gather about their surroundings before making decisions. Expanding the attention radius may improve decision accuracy and minimize environment uncertainty but also increases computational demands, making it impractical for large-scale or dynamic environments. Conversely, overly restrictive observations may lead to suboptimal or myopic decision-making. Finding an optimal tradeoff between situational awareness and resource efficiency is crucial for designing effective multi-agent strategies under uncertainty.

In this paper, we propose an algorithm that integrates Stackelberg games with an adaptive attention radius, allowing agents to dynamically adjust their level of awareness based on situational demands. By optimizing observation and decision-making simultaneously, our approach enhances coordination and efficiency in multi-agent environments while minimizing unnecessary resource usage. Through theoretical analysis and empirical validation, we demonstrate how adaptive observation can improve system performance and enable more scalable decision-making in uncertain environments. 

\section{Related Work}
    \subsection{Stackelberg Games}
Stackelberg games constitute a category of strategic games within game theory wherein players execute decisions sequentially rather than concurrently. In such games, one participant, designated as the leader, initiates by establishing their strategy, while the other participant, the follower, observes the leader's decision and subsequently determines their own strategy. This configuration enables the leader to anticipate the follower's response, thereby allowing the leader to optimize their strategy based on this foresight. Stackelberg games find common application in modeling scenarios such as smart grids \cite{maharjan2013dependable}, adversarial prediction problems \cite{liu2009game}, and price modeling \cite{tao2021stackelberg}. The capability of agents to make decisions in a sequential order renders the games dynamic and strategic, requiring players to foresee not only the immediate outcomes of their actions but also how future players will react to those actions, potentially modifying their strategies over time. 

Depending on the context of the game, the decision-making order may not be immediately evident. In collaborative games where agents can communicate, the ordering is subject to change over time. In \cite{hu2024playsfirstoptimizingorder}, the sequence of agents is determined at each time step. Rather than employing a naive methodology of testing all permutations and identifying the optimal, the authors propose an algorithm leveraging the Branch \& Bound concept \cite{mitten1970branch} to efficiently ascertain the optimal permutation, thereby facilitating decision-making based on that Stackelberg game. 


While this methodology has produced noteworthy results, it is essential to consider its practicality, particularly in scenarios with limited computational resources. In the referenced studies, the entire state space was continually observed, which might not be feasible when extending to large systems, such as a fleet of vehicles. Consequently, devising a strategy to diminish demand while sustaining interaction strategies is crucial. Additionally, the referenced studies depict environments where full coordination among all agents is possible, which might not be applicable in designing real-world scenarios.

\subsection{Vehicle Trajectory Estimation}




In the design of autonomous vehicles, a significant challenge is determining how to effectively interact with other vehicles occupying the same space. Contemporary designs commonly feature mechanisms for predicting the future trajectories of nearby vehicles \cite{huang2022survey} and pedestrians \cite{rudenko2020human}. 

The various methodologies for devising a trajectory estimation strategy are detailed in \cite{manas2025knowledgeintegrationstrategiesautonomous} and can be classified as either model-based or data-driven. A model-based approach \cite{xie2017vehicle} employs physical equations to describe the dynamics, potentially yielding high accuracy. Nevertheless, such models can be computationally burdensome. Conversely, a data-driven approach leverages past environmental explorations to formulate future decision-making policies. Although this method can be computationally efficient, it demands considerable time and resources for model training and may often fail to encompass all potential edge cases that might arise in future scenarios. Examples of this include reinforcement learning (RL) \cite{yuan2024evolutionary} and retrieval-augmented generation (RAG) \cite{cai2024driving}. A common compromise is to implement a hybrid model \cite{li2023physical}, which incorporates physical constraints into data-driven learning algorithms. 

\subsection{Resource-Aware Decision Making}
Selecting a decision-making policy for an autonomous agent in a MAS environment requires a trade-off between many different priorities such as optimality, safety, and resource consumption. With the rise of edge technologies applicable to robotics~\cite{tahir2025edge} and artificial intelligence (AI)~\cite{mcenroe2022survey} algorithms requiring large amounts of data, resource consumption has necessarily become a bigger focus of which algorithm to use.

A popular way to model resource-constrained systems and how they are able to make decisions is partially observable Markov decision processes (POMDPs)~\cite{kurniawati2022partially}. In this model, the visual capabilities of each agent are constrained, leading to incomplete knowledge of the surrounding environment. The extent to which observations can be made is dependent on the computational budget that each agent has, in addition to the physical limitations of any sensors they are using to make observations. This type of model also takes into consideration the stochastic nature of the environment and uses a Bayesian approach to make the best decision with the knowledge that is had. 

The topic of resource-constrained decision making for robotics systems has been covered in past works of research~\cite{alcorn2025situational}, which highlights the significance of this issue in robotics. Therefore, continuing to study this topic is of high importance and will have significant impacts on future autonomous technologies. 

\section{Problem Formulation}
\label{sec:formulation}
    \begin{table}[]
    \centering
    \begin{tabular}{|c|l|}
        \hline
        $v_i$ & Agent $i$ \\
        $\mathbf{x}^t$ & State at time $t$ \\
        $\mathbf{u}^t$ & Control inputs at time $t$ \\
        $\gamma^i$ & Strategy for agent $i$ \\
        $J$ & Cost of a chosen strategy \\
        $\gamma^*$ & Strategy following Stackelberg equilibrium \\
        $r^{att}$ & Attention Radius \\
        $r^{obs}$ & Observation Radius \\
        $\gamma^{opt}$ & Optimal Strategy (no Stackelberg) \\
        $o_i^t$ & Observation taken by agent $i$ at time $t$ \\
        $\sigma$ & Uncertainty \\
        \hline
    \end{tabular}
    \caption{List of Variables}
    \label{tab:list-of-variables}
\end{table}

In the proposed approach, we present a comprehensive framework designed to enable resource-optimized interactions among multiple agents by integrating three core components: estimation, observability, and decision making. Estimation allows each agent to infer critical information about its environment and the state of other agents. Observability ensures that the system can monitor key variables and detect changes or anomalies that may impact performance or coordination. Finally, decision making leverages the information gathered through estimation and observability to guide agents toward actions that collectively optimize the use of shared resources. By combining these elements, our approach promotes efficient, scalable, and intelligent cooperation in complex multi-agent systems.

The contribution of this work lies in the introduction of an adaptive attention radius, which serves as a novel mechanism for capturing both uncertainty and risk in dynamic environments. This adaptive mechanism enables agents to adjust the scope of their attention based on contextual factors, such as the level of uncertainty or the presence of potential threats, allowing for more informed and flexible decision making. Importantly, the adaptive attention radius also facilitates a dynamic balance between relying on optimal, low-complexity strategies and switching to more resource-intensive, adaptive strategies when the situation demands it. This flexibility ensures that the system remains both efficient and responsive, optimizing performance while minimizing unnecessary computational overhead.

\subsection{Strategy and Cost} 
\label{subsec:strategy}
Consider a set of $N$ autonomous agents $\mathbf{v}_N = (v_1, ..., v_N)$. At time $t$, each agent $v_i$ has a state $x_i^t \in \mathbb{R}^n$ and control input $u_i^t \in \mathbb{R}^m$. The dynamics are represented as a sequential trajectory game with the governing equation

\begin{equation}\label{eqn:governing-equation}
    \mathbf{x}^{t+1} = f(\mathbf{x}^t,\mathbf{u}^t),
\end{equation}
where $\mathbf{x}^t = (x_1^t, ..., x_N^t)$ and $\mathbf{u}^t = (u_1^t, ..., u_N^t)$.

The control inputs $u_i^t$ for the set of future time steps $(t, ..., t+T-1)$ are determined by a strategy $\gamma_i : \mathbb{R}^m \times [0,T] \rightarrow \mathcal{U}^i$. The implementation of a strategy $\gamma$ results in the future time steps $(t, ..., t+T-1)$ to take states $(x_i^t, ..., x_i^{t+T-1})$ with the resulting state cost functions of $(g(x_i^t, u_i^t), ..., g(x_i^{t+T-1},u_i^{t+T-1}))$. Therefore, the cost of the strategy $\gamma^i$ is defined as 

\begin{equation}\label{eqn:individual-strategy-cost}
    J(\gamma^i) = \sum\limits_{k=0}^T g(x_i^{t+k},u_i^{t+k}).
\end{equation}

The set of strategies for all $N$ agents $\boldsymbol{\gamma} := (\gamma_1,..., \gamma_N)$ has a total cost represented by

\begin{equation}\label{eqn:total-strategy-cost}
    \mathbf{J}(\boldsymbol{\gamma}) = \sum\limits_{i = 1}^N J(\gamma^i).
\end{equation}

The optimal joint strategy $\gamma^* = (\gamma_1^*, ..., \gamma_N^*)$ is the global Stackelberg equilibrium at time $t$, which means that for all $i = (1, ..., N)$

\begin{multline}\label{eqn:stackelberg-equilibrium}
    \sup_{\gamma^{>i} \in R^{>i}(\gamma^{\leq i, *})} J(\gamma^{\leq i, *}, \gamma^{>i}) \\ \leq \sup_{\widetilde{\gamma}^{>i} \in R^{>i}(\gamma^{< i, *}, \widetilde{\gamma}^i)} J(\gamma^{<i, *}, \widetilde{\gamma}^i, \gamma^{>i}),
\end{multline}
for all $\widetilde{\gamma}^i$, where $R^{>i}$ is the optimal response map of agent $v_i$. 

\subsection{Estimation and Non-Coordinated Interactions}
In an environment where $N$ agents are coordinating with each other and one agent is acting independently, we represent the state of the independent agent at time $t$ as $\bar{x}^t$. Since the agent acts on its own, we want to be able to predict the future trajectory of the agent in the upcoming $T$ time steps. This future behavior will be represented as the vector $\mathbf{\hat{\bar{x}}_T^t} = (\hat{\bar{x}}^t, ..., \hat{\bar{x}}^{t+T})$, where $\hat{\bar{x}}^t$ represents the estimated position of the agent at time $t$. 

Using the predicted future trajectory, a strategy can be built to determine the best strategy for the $N$ agents to work around the non-cooperative agent. This is done by assuming the non-cooperative agent is the leader in the Stackelberg game while all other agents will follow. In other words, the non-cooperative agent will be allowed to make their decision first, and all other agents will determine their best response to achieve their objectives. 

While the general case is that there are $N$ cooperative agents interacting with a single non-cooperative agent, it is possible or even likely that some of the agents do not need to be involved in the joint decision making due to being a sufficient distance removed from this interaction. Agents that are not involved in the multi-agent interaction will follow a predetermined optimal strategy $\gamma^{opt}$. To determine whether an agent will take place in the interaction, we define an attention radius, $r_i^{att}$, to be the distance at which agent $v_i$ is able to take in information about surrounding vehicles and obstacles. This means that if the distance between agent $v_i$ and any other agent is greater than $r_i^{att}$, they will not take place in the interaction process and will continue following $\gamma^{opt}$.

\subsection{Observations}
\label{subsec:observations}
\begin{figure}
    \centering
    \includegraphics[width=0.8\linewidth]{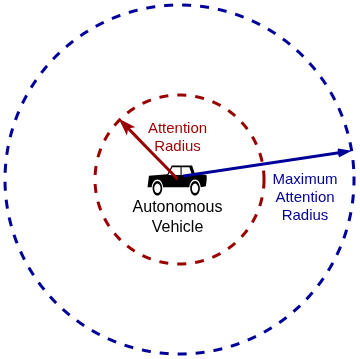}
    \caption{Visual representation of attention radius compared to the maximum distance that the vehicle can detect objects}
    \label{fig:observation-radius}
\end{figure}
An agent $v_i$ takes observation $o_i^t$, which is a partial representation of the actual state $x_i^t$. The observation radius, $r_i^{obs}$ is defined as the distance at which the agent is able to take in information about surrounding vehicles and obstacles. This value is fixed and is dependent on the hardware used in constructing the agent. We assume that the vehicle is able to see anything within $r_i^{obs}$ with sufficient accuracy as long as the line of sight is unobstructed. 

Additionally, we define the attention radius $r_i^{att}$. All agents $v_i \in \mathbf{v}_N$ will take actions that follow the optimal path to their destination unless there is another agent or obstacle detected within $r_i^{att}$. Therefore, our objective is to set $r_i^{att}$ such that resource usage can be minimized while the optimal strategy to avoid a collision can still be found. Specifically the radius becomes the solution to the optimization problem
\begin{equation}
\begin{aligned}\label{eqn:resource-optimization}
    r^{att}_i = \argmin_{r} \quad & J(\gamma_i, r, \sigma, n) \\
    \mathrm{s.t.} \quad & \gamma_i = \gamma_i^* \\
    & R_{min} \leq r \leq R_{max} \\
    & \sigma \leq \Sigma_{max} \\
    & n = N,
\end{aligned}
\end{equation}
where $J(\cdot)$ represents the cost function including the resources used and $\sigma$ is the risk associated with collisions and uncertainty in the environment. 

\subsection{Convergence}
In Equation \ref{eqn:resource-optimization}, we propose an optimization problem for finding the attention radius at each time step. In this section, we aim to prove that the solution to this problem is unique and will converge under specific conditions. 

\subsubsection{Convergence via Gradient Methods}
To show the rate of convergence for the optimal attention radius at each time step, we introduce Theorem \ref{thm:converge-rate}.

\begin{theorem}\cite{nesterov2013introductory}\label{thm:converge-rate}
    Let $f : \mathbb{R}^n \rightarrow \mathbb{R}$ be $L$-smooth, lower-bounded by $f^*$, and satisfy the Polyak-Łojasiewicz (PŁ) condition with PŁ constant $\mu > 0$. Then, the sequence $\{x_t\}$ of iterates produced by gradient descent instantiated with $0 < \eta \leq \frac{1}{L}$ converges to some point $x_* \in \mathbb{R}^n$, with 

    \begin{equation}
        ||x_t - x_*||_2^2 \leq \frac{8\eta L^2}{\mu^2}\left(1 - \frac{\mu}{L}\right)^{t-1}\left(f(x_0) - f_*\right).
    \end{equation}
\end{theorem}
Using Theorem \ref{thm:converge-rate}, since $\left(1 - \frac{\mu}{L}\right) < 1$, we can see that the $L$-2 distance between $x_t$ and $x_*$ converges to zero at an exponential rate, making gradient descent an efficient method for finding the optimal attention radius for each time step. 

To allow for this convergence guarantee to apply, the cost function $J(\cdot)$ must be designed such that it meets the criteria outlined in Theorem \ref{thm:converge-rate} ($L$-smooth function that satisfies the PŁ condition). 

\subsubsection{Uniqueness of Solution}
As highlighted previously, convergence is guaranteed under the specified conditions, but we also must prove that the solution is unique under those same conditions. Thus, we introduce Theorem \ref{thm:uniqueness} below to guarantee uniqueness.

\begin{theorem}\cite{ben2001lectures}\label{thm:uniqueness}
    Consider an optimization problem

    \begin{equation}
    \begin{aligned}
        \min \quad & f(x) \\
        \text{s.t.} \quad & x \in \Omega,
    \end{aligned}
    \end{equation}

    where $f : \mathbb{R}^n \rightarrow \mathbb{R}$ is strictly convex on $\Omega$ and $\Omega$ is a convex set. Then the optimal solution must be unique. 
\end{theorem}

Therefore, we have shown that a strictly convex cost function that is $L$-smooth and satisfies the PŁ condition will converge to the unique solution exponentially quickly.

\section{Method}
\label{sec:method}
    \subsection{Algorithm}
\begin{figure}
    \centering
    \includegraphics[width=0.85\linewidth]{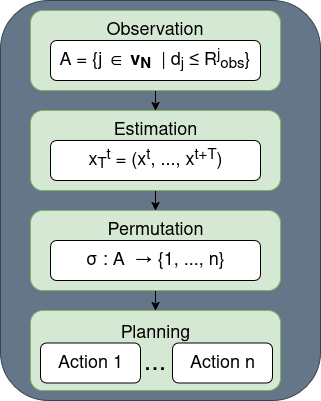}
    \caption{Visual diagram of the flow of logic in the decision making among collaborative agents}
    \label{fig:decision-making-flow}
\end{figure}

The objective of the MAS agents $\mathbf{v}_N$ is to build an optimal strategy that allows for all agents to achieve their respective objectives with minimal joint cost, as calculated in Equation \ref{eqn:total-strategy-cost}. However, achieving this optimal outcome can be impractical due to the resources and time required to find this joint strategy. The complexity of building a strategy is largely dependent on the attention radius $r^{att}$, since a large radius will require observing larger parts of the environment to attempt to avoid other agents, obstacles, etc. Therefore, it can be efficient to restrict the radius when necessary to eliminate unnecessary waste of resources. To do this, we propose the dynamic attention radius for resource-constrained multi-agent systems (DARRMS) algorithm to set an appropriate attention radius based on the observed risk and uncertainty in the environment. 

As outlined in Algorithm \ref{alg:strategy-selection}, an agent at time $t$ will observe within their attention radius and determine if there are any non-collaborative agents present. If there are, a prediction for the non-collaborative agent, $\mathbf{\hat{\bar{x}}}_T^t$, is calculated for the next $T$ time steps. Then, the uncertainty level $\sigma$ of the environment is calculated, followed by determining the optimal joint strategy of all agents that are interacting with the non-collaborative agent. Finally, the radius to be used for the next $T$ time steps is calculated by solving the optimization problem defined in Equation \ref{eqn:resource-optimization}. 

\begin{algorithm}
\caption{Decide between taking optimal strategy or collision avoidance}\label{alg:strategy-selection}

\begin{algorithmic}
    \State \textbf{Define}
    \State $T$ 
    \State \textbf{Initialize}
    \State $r_i = R_{init}$
    \State $x_i = x_{init}$
    \For{$t = (0, T, 2T, ...)$}
        \State $\widetilde{x} = \textit{observe-agent}$
        \If{$||x_i - \widetilde{x}||_2 \leq r_i$}
            \State $\widetilde{x}_{est}^T = \textit{predict-trajectory}$
            \State $\sigma = \textit{measure-uncertainty}(r_i, \widetilde{x}_{est}^T)$
            \State $\gamma = \gamma_i^*$
            \State $r_i = \textit{solve-opt}(\sigma, \gamma_i, \widetilde{x}_{est}^T)$
        \Else 
            \State $\sigma = \textit{measure-uncertainty}(r_i)$
            \State $\gamma_i = \gamma_{opt}$
            \State $r_i = \textit{solve-opt}(\sigma, \gamma_i)$
        \EndIf
    \EndFor
\end{algorithmic}
\end{algorithm}

If a non-collaborative agent is not detected within the attention radius, then the strategy followed is a predetermined optimal strategy. The value of $r^{att}$ is then calculated for the next iteration using the environment uncertainty and planned trajectory.
A visual representation of the order of decision making is shown in Figure \ref{fig:decision-making-flow} along with the mathematical formulation of each step in the algorithm. The stages of decision making are reduced to observation, estimation, permutation, and planning. Observation is where each agent looks within their attention radius to see which agents are present. Then, if the non-collaborative agent is present within $r^{att}$, estimation comes next where the next $T$ time steps of their trajectory are predicted. After estimation, the permutation stage determines the order in which the agents that are collaborating with each other will make decisions. Finally, the agents will make decisions in order and take an action. 



\section{Simulation and Results}
    \subsection{Environment}
\label{subsec:environment}
To test this algorithm, we set up a testing environment using an online simulator that models autonomous vehicles moving in a 2-D environment. Using this environment, we are able to implement the DARRMS algorithm and study its effectiveness. The simulation was set up with four agents, where three are collaborative and one is acting independently. Each agent has an objective of reaching their own destination, and the collaborative agents have a joint cost function that allows them to consider how their actions affect others. All agents in this simulated environment utilize the dynamics defined by the equation
\begin{equation}
    \begin{bmatrix}
        \dot{x} \\
        \dot{y} \\
        \dot{v} \\
        \dot{\theta}
    \end{bmatrix} = \begin{bmatrix}
        0 & 0 & \cos(\theta) & 0 \\
        0 & 0 & \sin(\theta) & 0 \\
        0 & 0 & 0 & 0 \\
        0 & 0 & 0 & 0
    \end{bmatrix}\begin{bmatrix}
        x \\
        y \\
        v \\
        \theta
    \end{bmatrix} + \begin{bmatrix}
        0 & 0 \\
        0 & 0 \\
        1 & 0 \\
        0 & 1 \\
    \end{bmatrix}\begin{bmatrix}
        a \\
        \omega
    \end{bmatrix},
\end{equation}
where $x, y$ are the coordinates in space, $\theta$ is the angle, $v$ is the linear velocity, $\omega$ is angular velocity, and $a$ is linear acceleration.

\begin{figure}
    \centering
    \includegraphics[width=\linewidth]{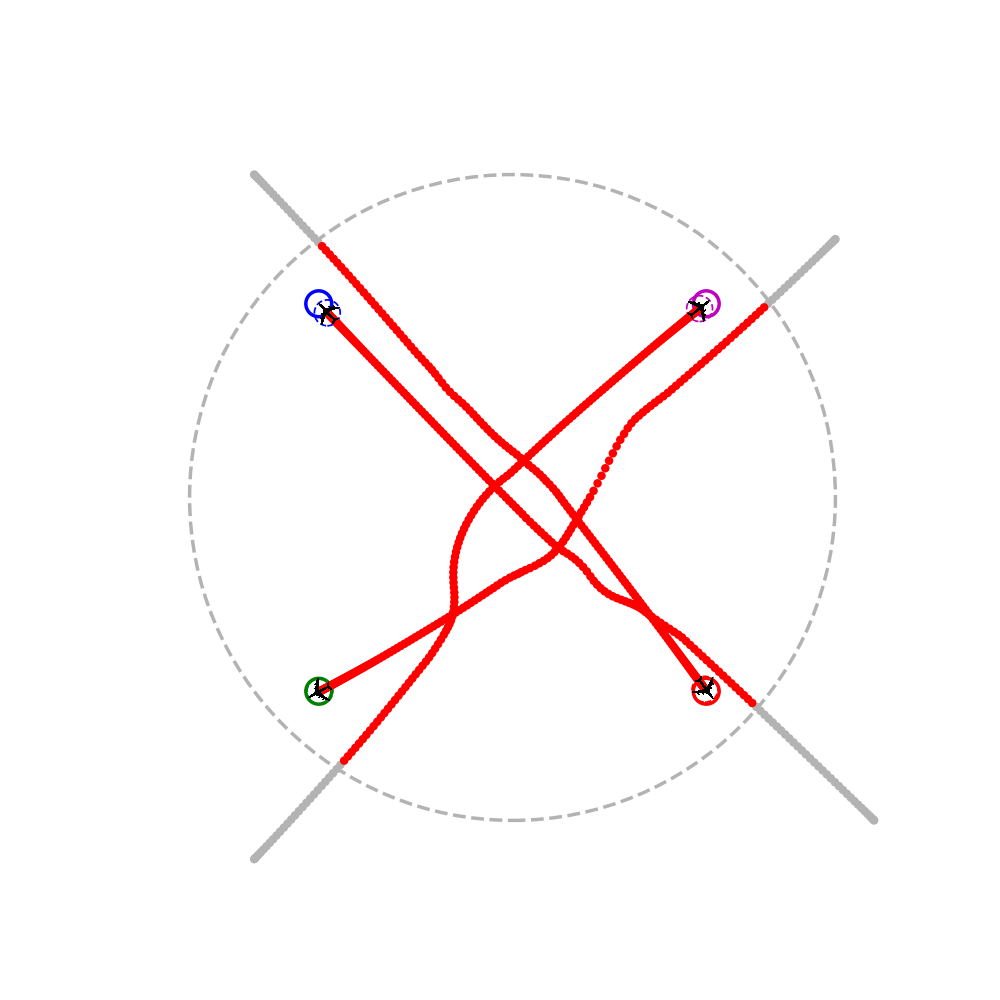}
    \caption{Results of simulated trajectories when implementing DARRMS algorithm. Here, there is a single non-collaborative agent (starts in top-left) while the others are collaborative}
    \label{fig:result-trajectories}
\end{figure}
\subsection{Simulation Setup}
We conducted two critical comparative analyses in the simulation environment to assess the feasibility of the proposed algorithm. Firstly, we evaluated the performance of the DARRMS algorithm against a baseline by analyzing the total time required for all agents to achieve their objective of reaching their respective destinations. Secondly, we examined the average attention radius throughout the duration of the simulation. The findings were derived from simulations executed across a substantial number of randomized environments.

A primary objective of the DARRMS algorithm is to diminish computational demand while preserving the integrity of other crucial performance metrics. In this context, two key performance indicators are the time to destination and the incidence of total collisions, thereby enabling the assessment of performance of the adaptive radius in DARRMS relative to the baseline of a fixed radius.

The attention radius exerts a significant influence on the resource consumption, as the size of the radius directly impacts the likelihood of using a resource efficient strategy vs. a costly strategy. How different strategies affect the consuption of computational resources is covered in Section. Hence, assessing the attention radius demonstrates the extent to which resource consumption is reduced for the implementation of the algorithm.

A symmetric environment was implemented before performing any analysis to determine if the objective of reaching the destination without any collisions was achieved for all agents. Figure \ref{fig:result-trajectories} shows the result of this, where all agents have reached their destinations and their trajectories are plotted along the path taken. The software provided outputs of a similar visual at each time step, which was used to verify that every time step had no collisions taking place. 


\begin{table}[]
    \centering
    \resizebox{\columnwidth}{!}{
     \begin{tabular}{||l|c|c|c||}
    \hline
      & Fixed Radius & DARRMS \\ \hline
     Average Observation Radius $(m)$ & 100.00 & \textbf{90.29} \\ \hline
     Average Time to Destination $(s)$ & \textbf{30.168} & 32.216 \\ \hline
     Average Resource Consumption Rate $(kB/s)$ & 473.6 & \textbf{220.8} \\ \hline
    \end{tabular}
    }
    \caption{Comparison of the average values of observation radius and time to destination for the fixed radius strategy and DARRMS algorithm where units are represented as unit distance}
    \label{tab:performance-comparison}
\end{table}

The study that followed was performed by running 500 iterations of the simulation environment while randomizing the starting positions and targets for all four agents in the environment. This was done to allow for variety and to prove that convergence and high performance existed under differing scenarios. Each unique environment was run with the baseline algorithm and DARRMS to produce results that allowed for a fair comparison of the algorithms. The results of this study are provided in Table \ref{tab:performance-comparison}, which provides the average metrics over all of the simulated environments for both algorithms. 

\subsection{Overview of Results}

The results provided provide insight on how the DARRMS algorithm compares to the baseline in the three performance metrics. As intended, DARRMS produces a lower average attention radius while the baseline has a fixed radius, since it does not allow for it to change over time. Comparing the two main performance metrics, time to destination and resource consumption rate, highlights the benefit of using the DARRMS algorithm. The time to destination is only marginally improved when using the fixed radius strategy compared to DARRMS, which is an expected cost due to the reduced forsight. However, the resource consumption is reduced by more than $50\%$ when using the DARRMS algorithm, which is a significant reduction. This shows the true value of this algorithm, and how a small cost in performance can bring significant reductions in the resource demands. 

The validity of these results is clear, showing that the resource consumption is dramatically reduced while maintaining similar levels of performance to the baseline, which fulfills the purpose of the DARRMS algorithm. 

\section{Conclusion}
    This paper presents a novel approach to multi-agent decision-making by integrating Stackelberg game theory with adaptive observation strategies. Our proposed algorithm addresses the critical challenge of balancing situational awareness with resource constraints, offering a dynamic method for agents to adjust their attention radius based on the evolving needs of the environment. Theoretical and empirical results demonstrate that our approach not only enhances coordination and efficiency but also provides a scalable solution for complex, resource-limited multi-agent systems. Future research could extend this framework to more complex interaction models or investigate its application in real-world scenarios, further validating its effectiveness and scalability.

\printbibliography

\end{document}